\documentclass[conference]{IEEEtran}
\IEEEoverridecommandlockouts
\usepackage{amsmath,amssymb,amsfonts}
\usepackage{algorithmic}
\usepackage{graphicx}
\usepackage{textcomp}
\usepackage{xcolor}
\def\BibTeX{{\rm B\kern-.05em{\sc i\kern-.025em b}\kern-.08em
    T\kern-.1667em\lower.7ex\hbox{E}\kern-.125emX}}

\usepackage[
backend=biber,
style=ieee,
]{biblatex} 
\usepackage[caption=false, font=footnotesize]{subfig}

\usepackage{booktabs,tabularx}
\usepackage{array}
\begin{document}

\title{Lightweight Relational Embedding in Task-Interpolated Few-Shot Networks for Enhanced Gastrointestinal Disease Classification\\
}
\author{\IEEEauthorblockN{1\textsuperscript{st} Xinliu Zhong}
\IEEEauthorblockA{\textit{Department of Biomedical Engineering} \\
\textit{National University of Singapore}\\
Singapore\\
xinliuzhong@u.nus.edu}

\and
\IEEEauthorblockN{2\textsuperscript{nd} Leo Hwa Liang\IEEEauthorrefmark{1}}
\IEEEauthorblockA{\textit{Department of Biomedical Engineering} \\
\textit{National University of Singapore}\\
Singapore\\
bielhl@nus.edu.sg}
\and
\IEEEauthorblockN{3\textsuperscript{rd} Angela S. Koh}
\IEEEauthorblockA{
\textit{National Heart Centre Singapore}; \\
\textit{Duke-NUS Medical School} \\
Singapore \\
angela.koh.s.m@singhealth.com.sg}
\and
\IEEEauthorblockN{4\textsuperscript{th} Yeo Si Yong\IEEEauthorrefmark{1}}
\IEEEauthorblockA{\textit{Lee Kong Chian School of Medicine} \\
\textit{Nanyang Technological University}\\
Singapore \\
siyong.yeo@ntu.edu.sg}
\and
\\
\\
\IEEEauthorrefmark{1}Corresponding authors

}


\maketitle

\begin{abstract}
Traditional diagnostic methods like colonoscopy are invasive yet critical tools necessary for accurately diagnosing colorectal cancer (CRC). Detection of CRC at early stages is crucial for increasing patient survival rates. However, colonoscopy is dependent on obtaining adequate and high-quality endoscopic images. Prolonged invasive procedures are inherently risky for patients, while suboptimal or insufficient images hamper diagnostic accuracy. These images, typically derived from video frames, often exhibit similar patterns, posing challenges in discrimination. To overcome these challenges, we propose a novel Deep Learning network built on a Few-Shot Learning architecture, which includes a tailored feature extractor, task interpolation, relational embedding, and a bi-level routing attention mechanism. The Few-Shot Learning paradigm enables our model to rapidly adapt to unseen fine-grained endoscopic image patterns, and the task interpolation augments the insufficient images artificially from varied instrument viewpoints. Our relational embedding approach discerns critical intra-image features and captures inter-image transitions between consecutive endoscopic frames, overcoming the limitations of Convolutional Neural Networks (CNNs). The integration of a light-weight attention mechanism ensures a concentrated analysis of pertinent image regions. By training on diverse datasets, the model's generalizability and robustness are notably improved for handling endoscopic images. Evaluated on Kvasir dataset, our model demonstrated superior performance, achieving an accuracy of 90.1\%, precision of 0.845, recall of 0.942, and an F1 score of 0.891. This surpasses current state-of-the-art methods, presenting a promising solution to the challenges of invasive colonoscopy by optimizing CRC detection through advanced image analysis.

\end{abstract}


\section{Introduction}
\begin{figure}[tbp]
\centerline{\includegraphics[width=0.5\textwidth]{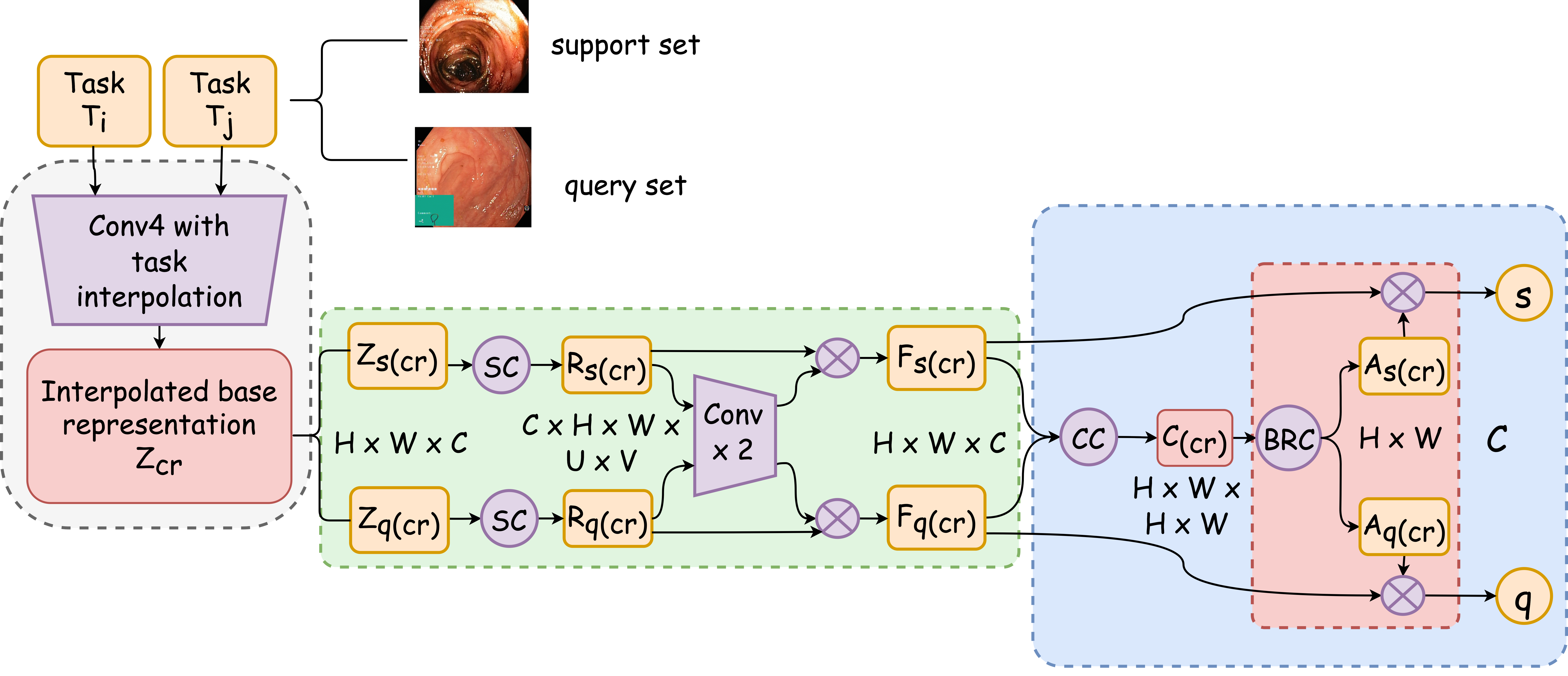}}
\caption{Overall architecture of our proposed model. Given tasks, $T$, including support and query set images are sampled as input into the Conv4 encoder. Within the encoder, task interpolation is performed to enhance the diversity of the task set at a randomly chosen layer. Then the extracted support and query features, $Z_{s(cr)}$ and $Z_{q(cr)}$, are updated by self relational module separately. These features, $F_{s(cr)}$ and $F_{q(cr)}$,  are integrated into the cross-relational module, which employs a lightweight bi-level routing attention mechanism to identify common areas of focus, $A_{s(cr)}$ and $A_{q(cr)}$, and derive the final outputs, $s$ and $q$.}
\label{fig:arct}
\end{figure}
Gastrointestinal (GI) diseases encompass a range of conditions that can significantly impact health. Among these, colorectal cancer (CRC) is a notable concern due to its high prevalence and mortality rate when diagnosed late. Early detection is key to improving prognosis, with the potential to prevent up to 90\% of deaths if identified promptly.

Endoscopic images are pivotal for diagnosing GI diseases, which predominantly arise from polyps in the lower digestive tract. The endoscope, a long tube with a fiber-optic camera, is used in this diagnostic procedure \cite{ramamurthy_novel_2022}. The surface morphological pattern of GI tract offers crucial clinical insights for surgical decisions and indicates disease aggressiveness.
Admittedly, colonoscopy has several strengths, but it's not without its limitations. The intricate and varied nature of the colon and polyps, combined with the static size of the structuring element in the morphological operator, complicates segmentation, especially when vessels around the polyps evolve along the periphery of the liver \cite{roy_automatic_2022}, leading to substantial background variation, fuzzy frames and noise
. Moreover, a multicenter prospective study \cite{lu2022real} has reported that up to 40\% of CRC cases with deep submucosal invasion were misdiagnosed as superficial invasive cancer, underscoring the influence of the complexity of assessment theory and the subjectivity among endoscopists in impeding the accurate diagnosis of GI diseases. 

Traditional computer vision techniques including Harris Corner Detection \cite{harris_combined_1988}, SURF \cite{surf}, and ORB \cite{orb}, are heavily dependent on domain-specific knowledge and often fall short in capturing intrinsic image features in gastroenterology. Consequently, there's increasing interest in artificial intelligence (AI)-based techniques for more precise and objective GI disease identification. Deep Learning (DL) has led to a shift towards Convolutional Neural Networks \cite{simonyan} (CNNs) like Conv4 and ResNet12, which automatically extract meaningful image features hierarchically, outperforming traditional methods in classification tasks. 

CNNs, foundational to various DL models, extract features through convolutions, influencing methods like LSTM, U-Net, and Inception. Dutta et al. \cite{dutta_efficient_2021} utilized a Tiny Darknet model, for efficient lesion detection. Another study \cite{zeng_image_2021} employed Xception \cite{chollet2017xception}, ResNet \cite{resnet}, and DenseNet \cite{huang2018densely} for ulcerative proctitis detection. Luo et al. \cite{luo_diagnosis_2022} combined CNNs and Recurrent Neural Networks (RNNs) to diagnose ulcerative colitis from endoscopic images, enhancing accuracy with a spatial attention module. In endoscopic medical imaging, the challenges are multifaceted, ranging from data constraints to inherent limitations of conventional DLs. Addressing this, our novel DL model is tailored for endoscopic images, with the following key attributes:
\begin{itemize}
    \item Traditional DL models, designed for generic images, falter with the specific textures and patterns of endoscopic images, especially when data is scarce in GI sector. By adopting a \textbf{few-shot learning} (FSL) paradigm, our model can \textbf{quickly understand and adapt} to the \textbf{unique characteristics} of endoscopic images, even with limited training data.
    
    \item Given the variability in endoscopic views due to different insertion angles and organ structures, over-fitting is a concern. Our model adds a \textbf{task interpolation} module in FSL to artificially diversify its training on various endoscopic perspectives, ensuring \textbf{robustness across different} endoscopic scenarios, overcoming the over-fitting issue.
    
    \item  Endoscopic procedures often capture a series of images. Our model does not just process each image in isolation; it\textbf{ understands the relationships and transition}s between consecutive endoscopic frames, offering a more comprehensive analysis.
    
    \item  In endoscopy, minute details can be critical for diagnosis. While traditional CNNs might miss out on such details, our model, with its \textbf{advanced light-weight bi-level routing attention mechanism}, zeroes in on regions of interest (ROIs) in endoscopic images, ensuring no detail is overlooked and eliminating unwanted pixels.
\end{itemize}



\section{Method}

In this article, we introduce a DL approach for endoscopic data that can solve specific problems like fine-grained image patterns, small dataset size, as well as view variations. The DL methodology includes a feature extractor, FSL, data augmentation, relational embedding, and bi-level routing attention.

Fig.\ref{fig:arct} delineates the architectural overview of our proposed model, encapsulating the methodologies detailed subsequently --- The rest of the section will be organized as follows: In Sec.\ref{sub:fe}, the rationale behind opting for Conv4 over ResNet12 for strategic feature extraction is elucidated. The ensuing section, Sec.\ref{sub:fsl}, articulates the incorporation of the Few-Shot Learning (FSL) paradigm for handling the limited-size endoscopic image data. Sec.\ref{sub:ti} explicates our approach to data augmentation through task interpolation, devised to synthesize multiple viewpoints. Moving forward, Sec.\ref{sub:re} describes the employment of both self- and cross-correlational embedding to discern relational representations within the CRC dataset, focusing on support and query images. Lastly, Sec.\ref{sub:bi} details the deployment of a lightweight bi-level routing attention mechanism, aimed at efficiently attending to common ROIs within the data.

\subsection{Feature Extractor}
\label{sub:fe}
To efficiently process the complexities of endoscopic images, we chose Conv4 as our main feature extractor due to its balance of simplicity and performance. Although Conv4 and ResNet12 are both viable CNN architectures, Conv4's effectiveness in handling the detailed, resource-intensive nature of endoscopic image analysis made it our preferred choice.

Conv4, a lightweight and computationally efficient CNN architecture, is particularly advantageous for FSL tasks, especially in scenarios with limited computational resources. Comprising four convolutional blocks, each with a convolutional layer, batch normalization layer, and ReLU activation layer, Conv4 effectively learns and extracts meaningful features from images through a hierarchical approach. Its simplicity and efficiency make it a suitable encoder for our FSL model in endoscopic image analysis, where data is often scarce and computational efficiency is crucial.

\subsection{Few-shot Learning}
\label{sub:fsl}
\begin{figure}[tbp]
    \centering
    \includegraphics[width=0.9\linewidth]{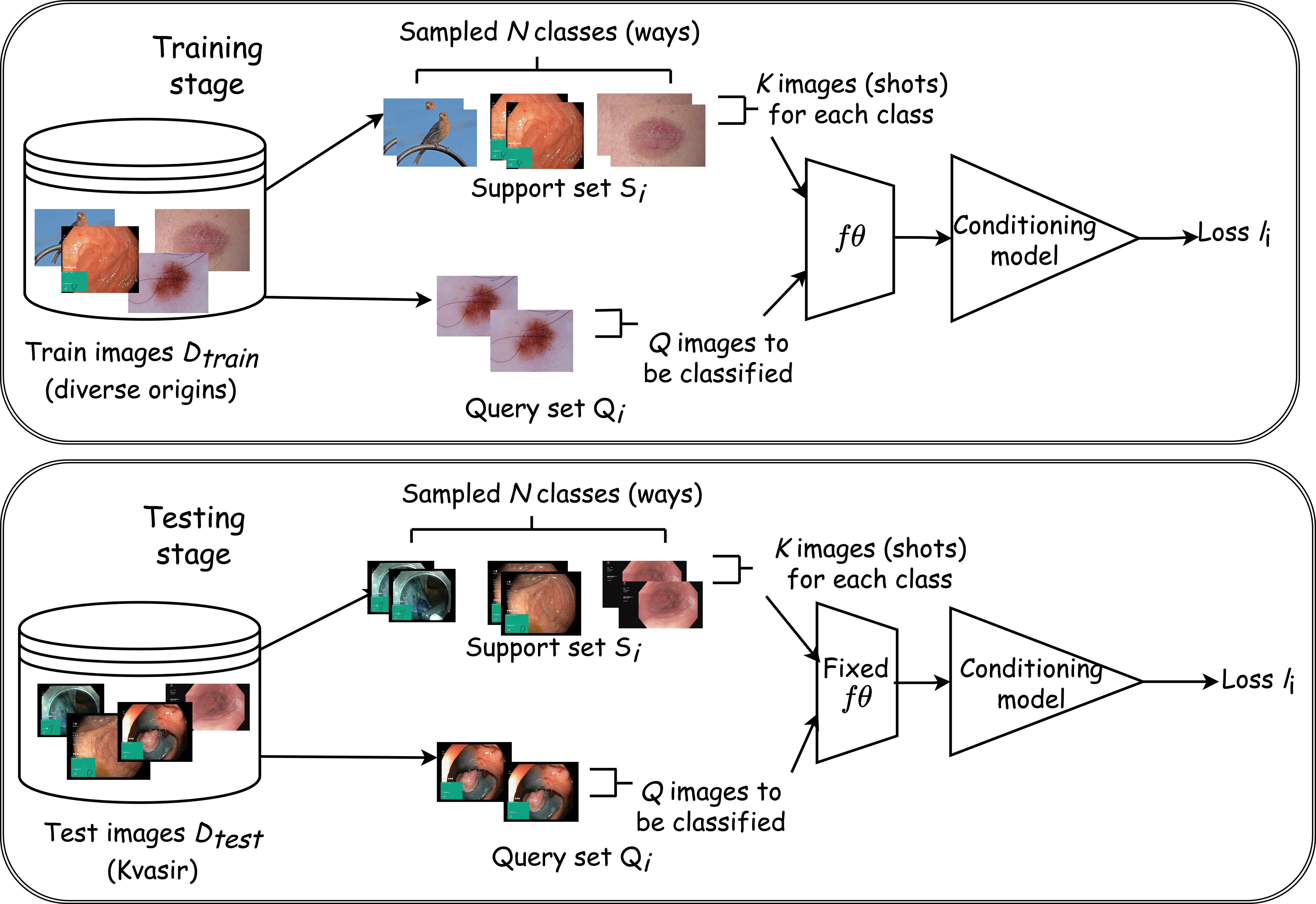}
    \caption{Adopted FSL paradigm for our approach. }
    \label{fig:fslp}
\end{figure}
FSL is a DL approach that is particularly advantageous for training datasets with scarce information, a common challenge in CRC image classification within the domain of endoscopy. We used this paradigm of training and testing in building our model, as shown in Fig.\ref{fig:fslp}. This approach swiftly learns new concepts from minimal data, providing a viable solution to the data scarcity issue prevalent in medical imaging, especially in endoscopy where procurement of labeled data can be particularly challenging in less-than-ideal imaging conditions. 

FSL builds a model by splitting the dataset $D$ into $D_{train}$ and $D_{test}$, ensuring the classes of the testing set should be unseen from the training set ($C_{train} \cap C_{test} = \varnothing$). Data from both sets are presented as tasks, each with a query set and a support set. For a task $T_{i}$, the support set $S_{i} = \{X^{j}_{s},y^{j}_{s}\}^{NK}_{j=1}$ includes $K$ images from each of $N$ classes, and the query set $Q_{i} = \{X^{j}_{q},y^{j}_{q}\}^{Q}_{j=1}$ comprises $Q$ images, forming an $N$-way-$K$-shot task.

\subsection{Data augmentation via task interpolation}
\label{sub:ti}
Data augmentation, pivotal in enhancing ML models, especially in endoscopic image analysis, addresses limited data availability. For FSL in endoscopy, task interpolation is employed to generate additional domains, thereby augmenting the training data. 
This technique seeks to learn invariant representations and enhance training consistency by densifying the task distribution in a cross-task manner. 

In a model \(f\) with \(L\) layers, the hidden representation of samples \(X\) at the \(l\)-th layer is \(H_l = f_{\theta_l}(X)\), with \(H_0 = X\) and \(L_s\) being the number of layers shared across all tasks. While meta-learning methods like MAML \cite{finn_model-agnostic_2017} share only a subset of the layers (i.e., \(L_s < L\)), metric-based methods like ProtoNet \cite{snell_prototypical_2017} share all layers (i.e., \(L_s = L\)). For a pair of tasks with their support and query sets sampled from the same label space, a layer \(l\) is randomly selected and task interpolation is applied separately on the hidden representations and corresponding labels of the support and query sets. The interpolated hidden representations and labels are \(\widetilde{H}_{s,l}^{cr} = \lambda H_{s,l}^i + (1 - \lambda)H_{s,l}^j\),
with \(\lambda \in [0, 1]\) sampled from a Beta distribution \( Beta(\alpha, \beta)\) and subscript 'cr' indicating 'cross'. Both the hidden representations of support and query set are replaced by the interpolated ones in our task interpolation. The resulting enlarged task set is utilized as input for our FSL model, enhancing its capability to discern and classify endoscopic images across changing endoscope viewpoints.


\subsection{Relational Embedding}
\label{sub:re}
\begin{figure}[htbp]
    \centering
    \includegraphics[width=0.9\linewidth]{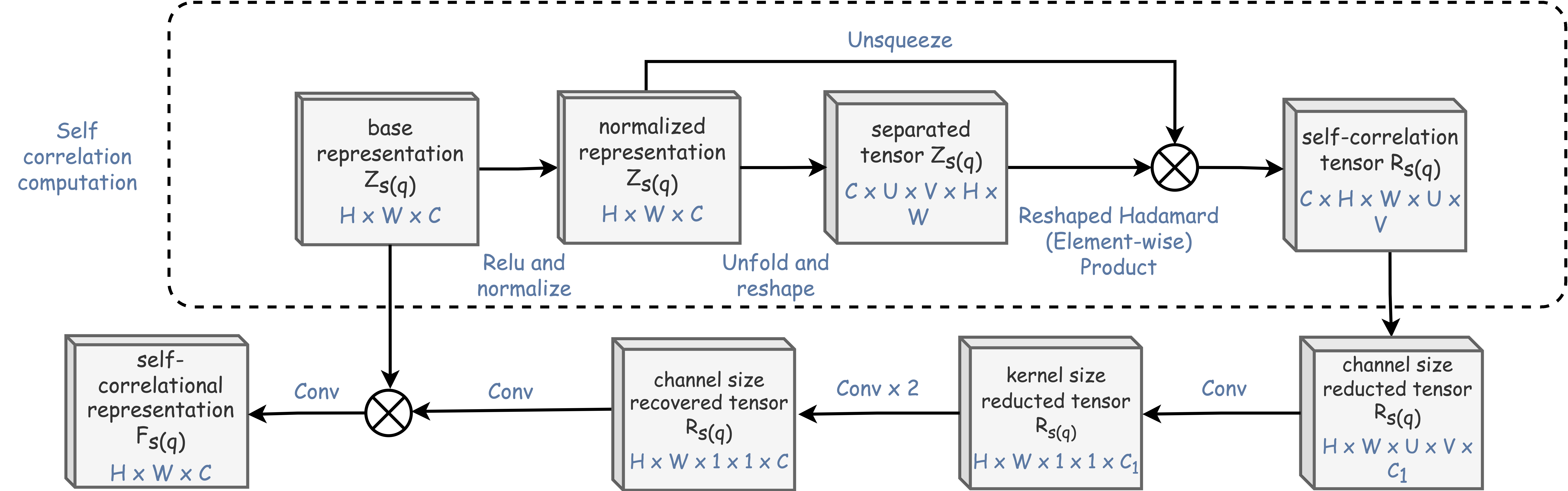}
    \caption{Structure of self-correlational representation module. Base representation $Z_{s(q)}$ undergoes self-correlation computation by computing its Hadamard product $R_{s(q)}$. After a series of convolutions, the self-correlational representation $F_{s(q)}$ is derived to attend to pertinent areas inside the image.}
    \label{fig:scr}
\end{figure}
Navigating through the intricacies of endoscopic images, especially within an FSL framework, demands a strategic approach to decipher and comprehend the subtle yet critical similarities and correlations among variables. This is particularly necessary in our target dataset, where the fine-grained medical images may appear superficially identical, and the data available is notably sparse. The challenge, therefore, is to unearth a relational structure that can simultaneously generalize across all embeddings, pinpointing ROIs that are crucial for precise classification and analysis.

The self-correlational representation (SCR) module \cite{renet}, depicted in Fig.\ref{fig:scr}, transforms the base representation $Z$ from our feature extractor to highlight ROIs in a single image, thereby preparing a reliable input to the cross-correlational attention (CCA) module for analyzing feature correlations between a pair of different images. Inside SCR, we calculate a Hadamard product $R$ within a neighbor window in a channel-wise manner as the self-correlation, followed by a series of convolutions, to produce the ultimate self-correlational feature. This feature, the same size as $Z$, is added back for reinforcement. 

The CCA module takes a query-support pair from SCR and produces corresponding attention maps $A_q$ and $A_s$, converting each representation to an embedding vector. It first computes the cross-correlation 4-dimensional tensor $C$ by transforming both query and support representations, $F_q$ and $F_s$, into more compact representations. Given potential unreliable correlations in $C$ due to appearance variations in FSL, a convolutional matching process refines $C$ into $C'$, from which co-attention maps $A_q$ and $A_s$ are derived, highlighting relevant content between query and support.

\subsection{Bi-level Routing Co-attention}
\label{sub:bi}
\begin{figure}[htbp]
    \centering
    \includegraphics[width=0.9\linewidth]{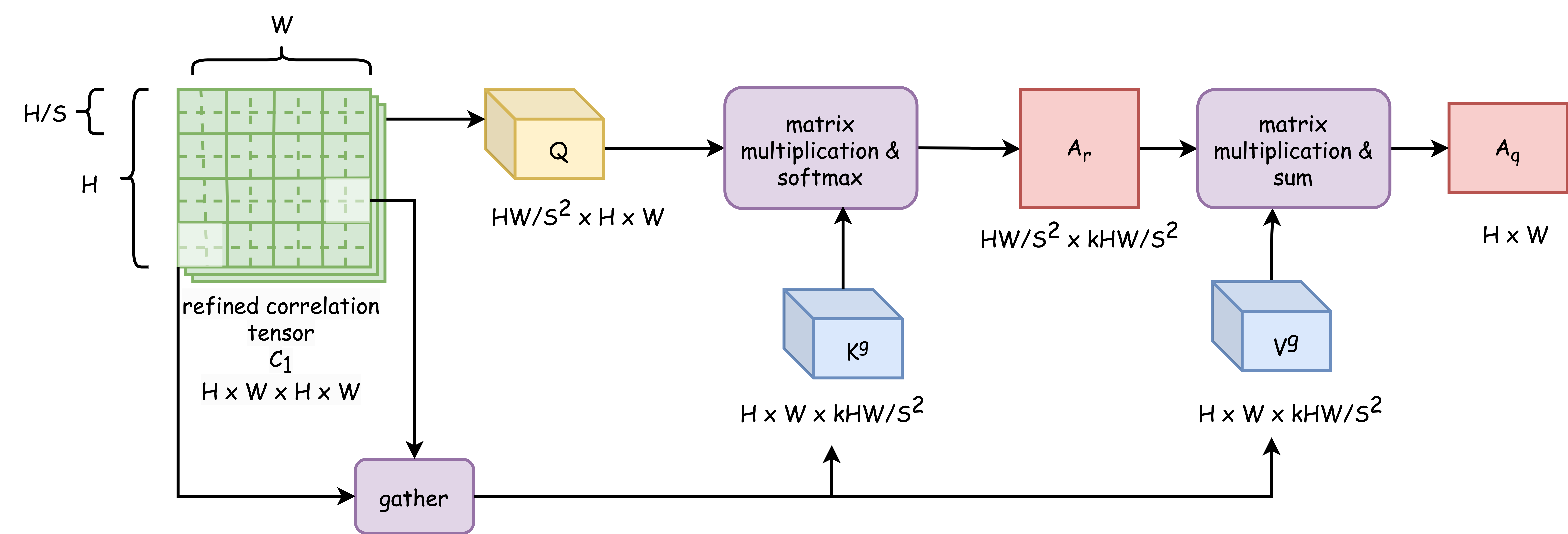}
    \caption{Structure of bi-level routing co-attention module for \(A_q\). The module simplifies attention calculations by gathering $topK$ key-value pairs $K^g$ and $V^g$ from coarse region-region affinity graph of the adjacency matrix $A^r$. Co-attention map for support set $A_s$ is similar.}
    \label{fig:biattn}
\end{figure}
The attention mechanism, initially from Nature Language Processing (NLP), shows its ability as an alternative to convolutions in Transformers \cite{vaswani_attention_2023}. This structure can deal with global features (long-ranged dependencies), which is necessary for endoscopic images of GI. Inside the CCA module, we adopted co-attention mapping operations to reveal cross relations between the query and support. 

In the standard Scaled Dot-product Attention calculation \cite{vaswani_attention_2023}, inputs include a query matrix $Q$, and matrices for keys $K$ and values $V$ of dimension $d_k$. The output is computed by applying a softmax function on the multiplication of the matrices as \(Attention(Q,K,V) = softmax(\frac{QK^T}{\sqrt{d_k}})V\).
Vision models often modify this by applying multi-head attention, concatenating parallel attention outputs for linear projection, with a complexity of $O(N^2)$ when there are $N$ pairs of key-value pairs for each query in $N$ queries.

To reduce the computational burden of attention, we implemented a bi-level routing attention mechanism \cite{zhu_biformer_2023}, illustrated in Fig.\ref{fig:biattn}. This method selects the most relevant key-value pairs from a coarse region-region affinity graph in the adjacency matrix $A^r$, created by multiplying region-level queries $Q^r$ and keys $K^r$. The routing index $I^r$ is obtained using a top-$k$ operation on $A^r$ for pruning. This technique maintains co-attention capability while reducing computation costs by optimizing the region size.

\section{Experiment}

\subsection{Datasets}
\begin{table}[h]
\caption{Summary of Datasets Used in the Study}
\centering
\begin{tabularx}{0.45\textwidth}{@{}lXXlX@{}} 
\toprule
\textbf{Dataset} & \textbf{\textit{Images}} & \textbf{\textit{Classes}}& \textbf{\textit{Purpose}} \\
\midrule
Kvasir-v2 \cite{kvasir} & 8000 & 8 & GI disease classification\\
Hyper-Kvasir \cite{borgli_hyperkvasir_2020} & 10,662 & 23 & GI disease classification\\
ISIC 2018 \cite{tschandl2018ham10000}\cite{codella2019skin} & 10,208 & 8 & Lesion classification \\
Cholec80 \cite{twinanda2016endonet} & 241,842 & 7 & Surgery tool recognition \\
Mini-ImageNet \cite{imagenet} & 60,000 & 100 & General object classification \\
\bottomrule
\end{tabularx}
\label{tab:dataset}
\end{table}
\begin{figure}[h]
\centering

\subfloat[
]{\includegraphics[width=0.11\textwidth]{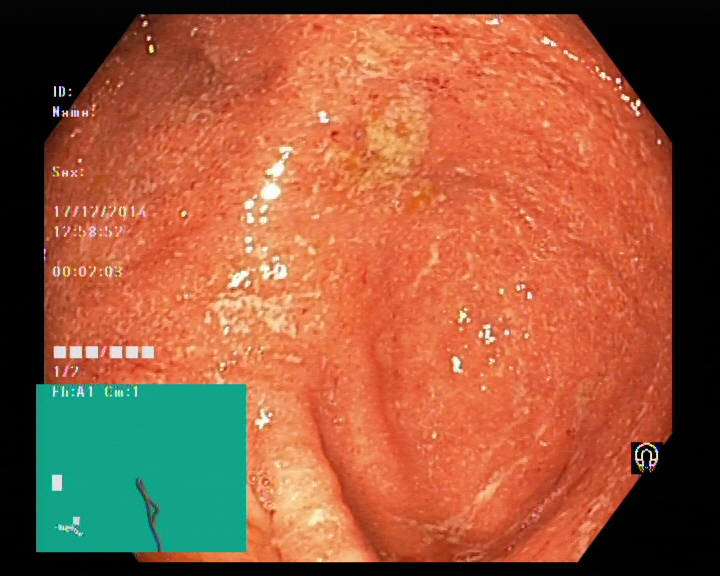}}
\hfill
\subfloat[
]{\includegraphics[width=0.155\textwidth]{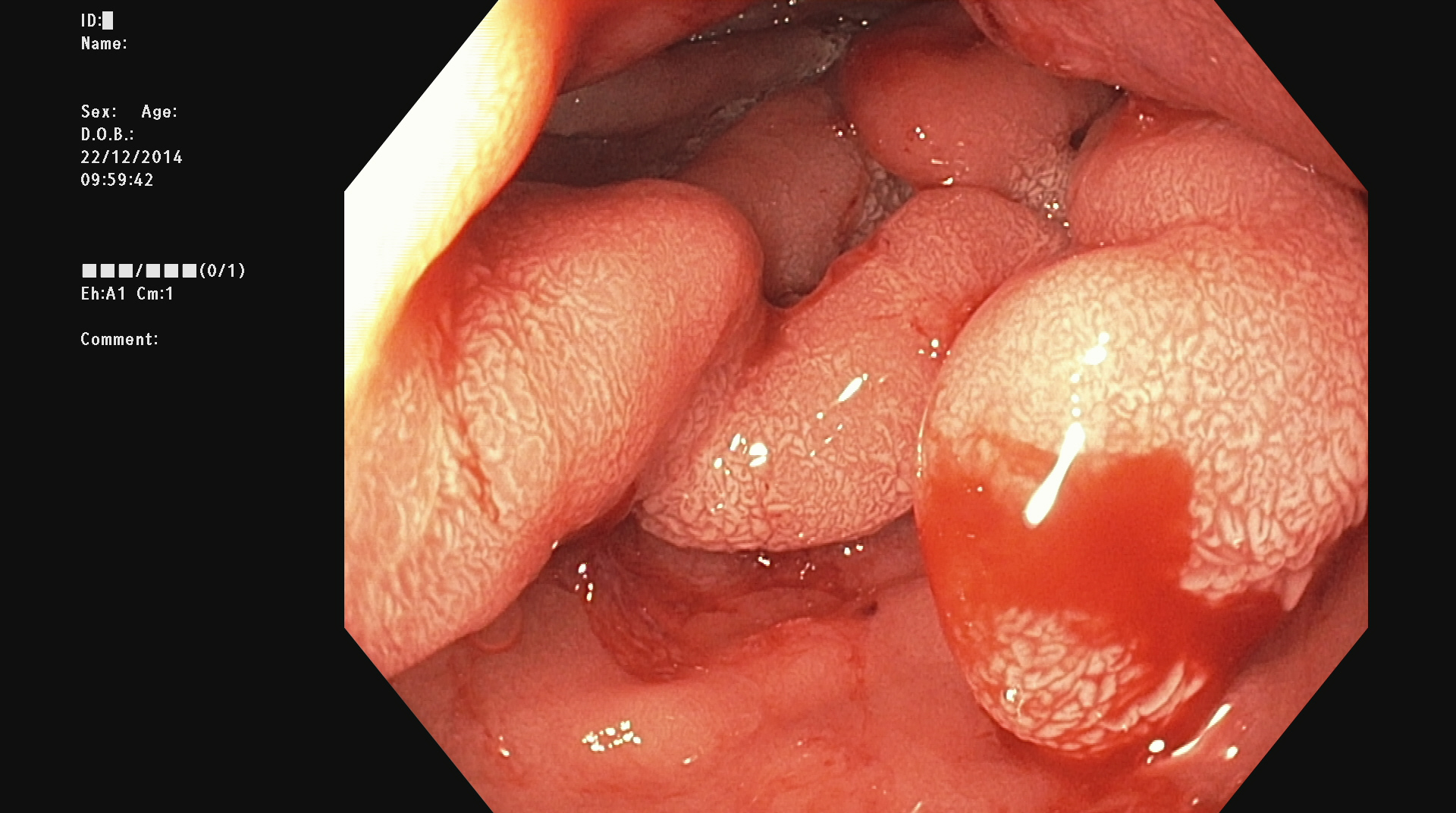}}
\hfill
\subfloat[
]{\includegraphics[width=0.11\textwidth]{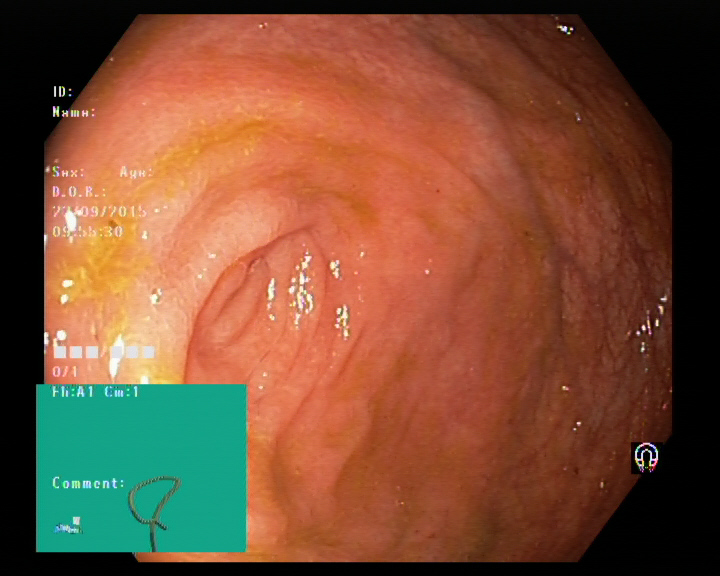}}
\hfill
\subfloat[
]{\includegraphics[width=0.11\textwidth]{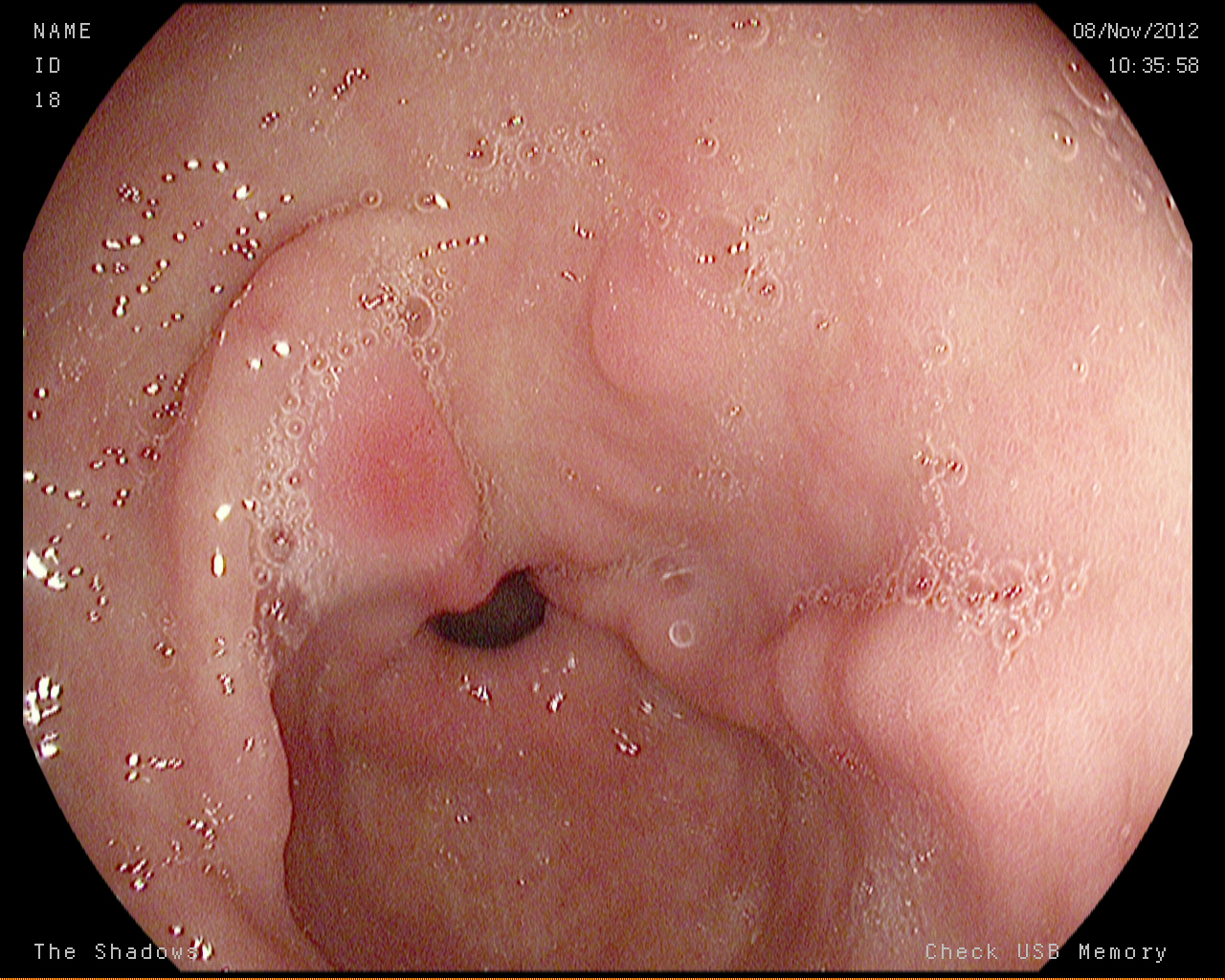}}
\\
\subfloat[
]{\includegraphics[width=0.115\textwidth]{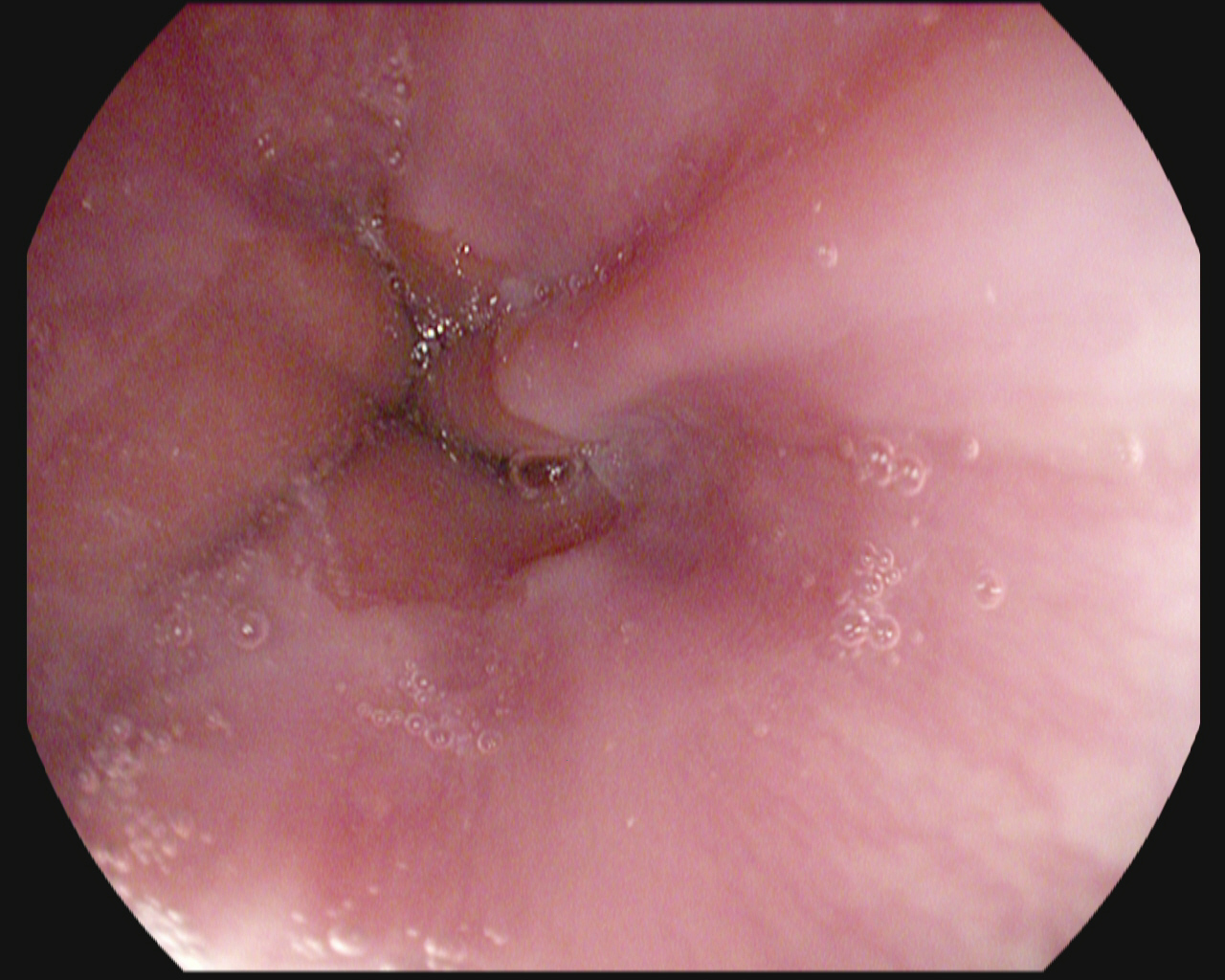}}
\hfill
\subfloat[
]{
\includegraphics[width=0.115\textwidth]{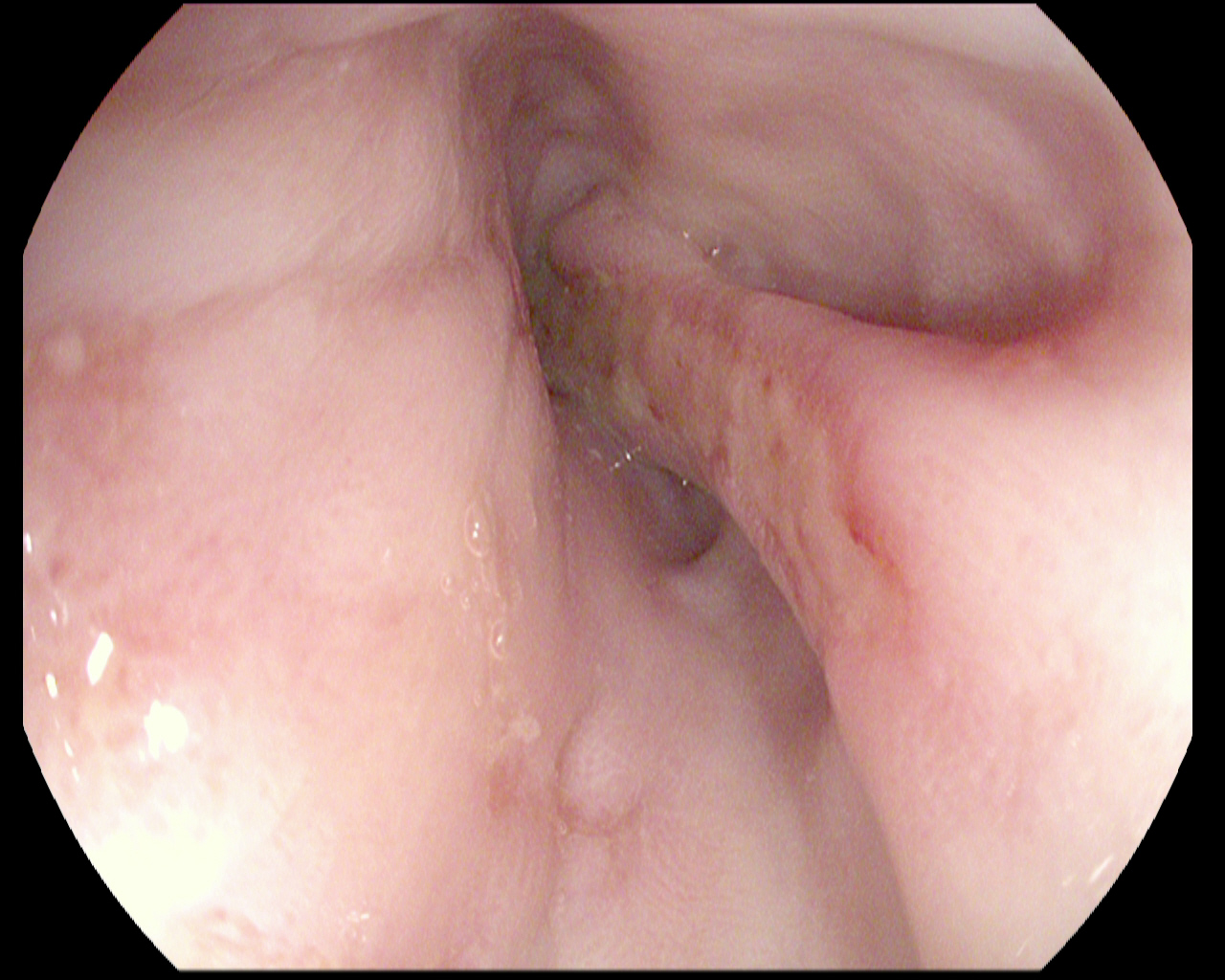}}
\hfill
\subfloat[
]{\includegraphics[width=0.115\textwidth]{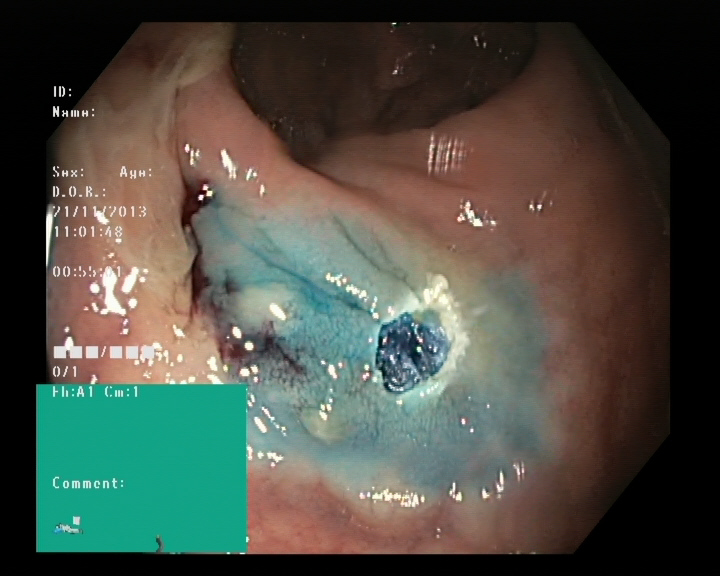}}
\hfill
\subfloat[
]{\includegraphics[width=0.115\textwidth]{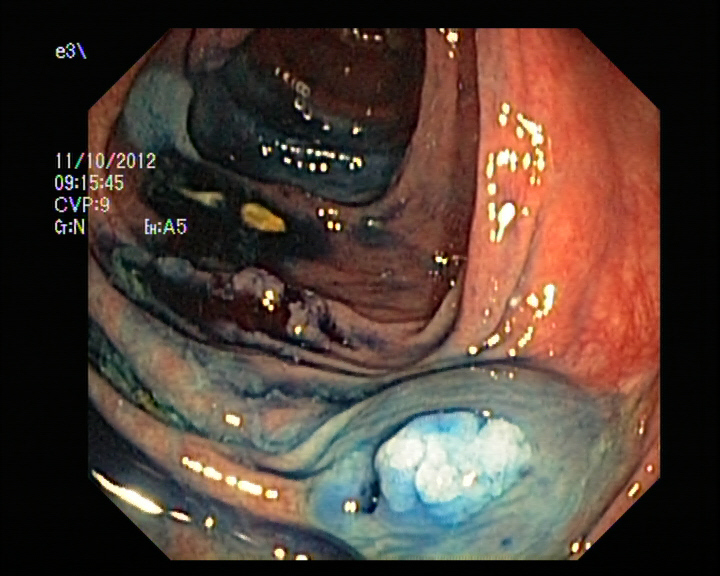}}

\caption{Examples of Kvasir dataset \cite{kvasir}. (a)-(h): ulcerative colitis, polyps, normal cecum, normal pylorus, normal z-line, esophagitis, dyed resection margins, dyed lifted polyps. \label{fig:kvasir}}
\end{figure}
To pretrain our model on endoscopic images, we utilized various datasets encompassing GI images, a broad-spectrum normal dataset, surgical tool images, and a skin disease dataset, enhancing the model's cross-domain generalization and reducing overfitting risk. The model underwent pretraining on ISIC 2018 \cite{tschandl2018ham10000}\cite{codella2019skin}, Cholec80 \cite{twinanda2016endonet}, and Mini-ImageNet datasets, followed by fine-tuning on the Hyper-Kvasir \cite{borgli_hyperkvasir_2020} dataset. This comprehensive fine-tuning involved 10,662 images across 23 GI disease classes, following the criteria in \cite{ramamurthy_novel_2022}. Incorporating a new full-connected layer head, the pretrained model was tailored for extracting endoscopic features. For testing, we employed the Kvasir-v2 dataset \cite{kvasir}, a specialized multi-class image dataset for computer-aided GI disease detection, featuring 8 classes with 1000 images each, shown in fig.\ref{fig:kvasir}. The datasets are summarized in Table.\ref{tab:dataset}.

\subsection{Implementation details}
Based on the configuration outlined in \cite{yao_meta-learning_2022}, we have eliminated the validation process. Additionally, we have adjusted our feature extractor, Conv4, by increasing the layer channels from 64 to 640, specifically to accommodate the relational embedding model structure. 
The experiments are performed under the N-way K-shot setting, where N = 2 for ISIC and N = 5 for the rest datasets, while K = 1 and number of query images is 15. The rationale Table.\ref{tab:para} of the experiment is listed below. 
\begin{table}[htbp]
    \caption{The Parameters of Experiment}
    \centering
    \begin{tabularx}{0.4\textwidth}{l c c}  
        \toprule
        \textbf{Type} & \textbf{\textit{Parameter}} & \textbf{\textit{Value}} \\
        \midrule
        Feature Extractor           &   Conv4 layer     &    640        \\
        \midrule
        Adjuster                    &   Optimizer       &    ADAM  \\
                                    &    Learning rate  &     0.0001(fixed) \\
                                    &  Weight decay     &    0.002    \\

        \midrule
        Data loader                 &    Batch size     &     128   \\
                                    &   Saving episodes &     500  \\
                                    &   Total episodes  &       5000 \\
        \midrule
        FSL setting            &  Training way     &       5\\
                                    & Testing way       &       2\\
                                    & Shot              &       1\\ 
        
        \bottomrule
    \end{tabularx}

    \label{tab:para}
\end{table}


In data pre-processing, the images are resized to $92\times92$, randomly cropped to $84\times84$ and horizontally flipped for augmentation, and then its pixels normalized with mean values of $[125.3, 123.0, 113.9]\setminus255.0$ and standard deviation values of $[63.0, 62.1, 66.7]\setminus255.0$ 
, inherited from ImageNet \cite{imagenet}. 


\begin{figure}[h]
\centering
\subfloat[
]{\includegraphics[width=0.16\textwidth]{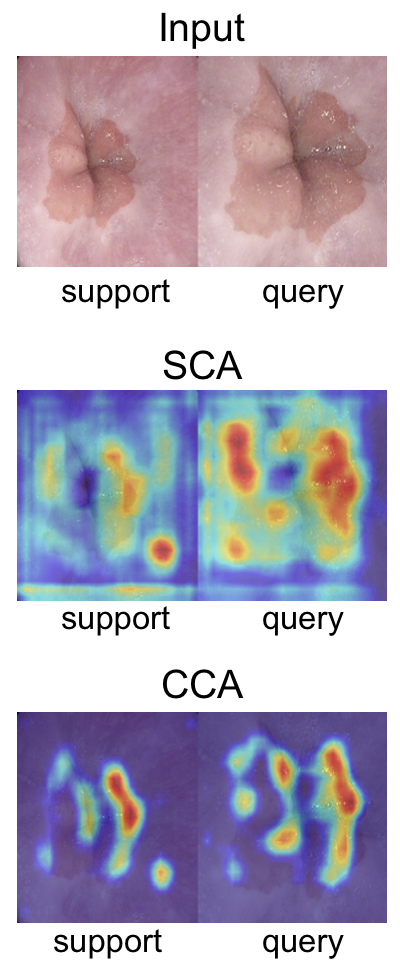}}
\hfill
\subfloat[
]{\includegraphics[width=0.16\textwidth]{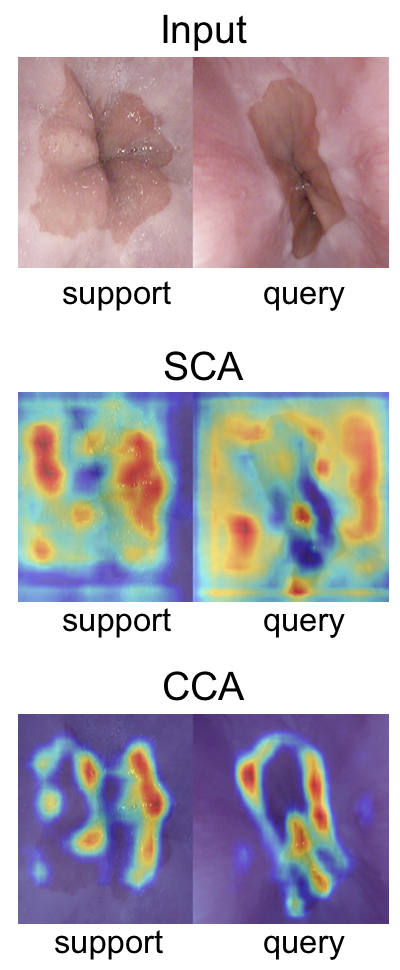}}
\hfill
\subfloat[
]{
\includegraphics[width=0.16\textwidth]{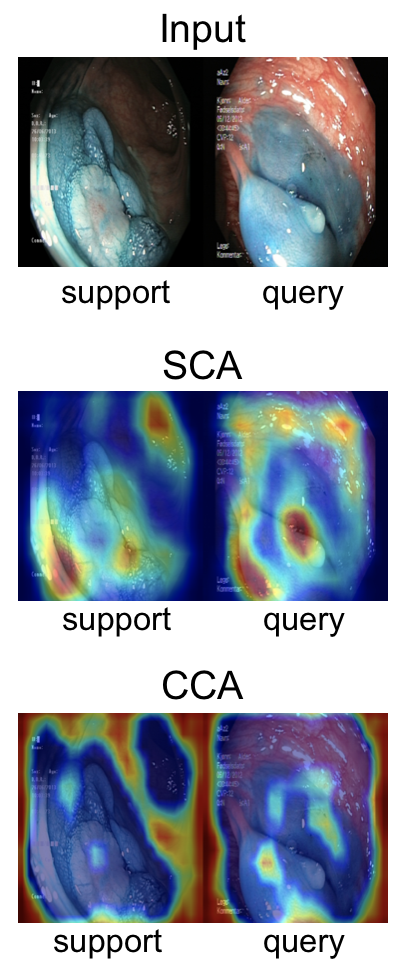}}
\caption{Attention Heatmaps of SCA and CCA Modules with different source of support (left) and query (right) images. (a) Standard and zoomed views of a normal Z-line. (b) Different images of the same normal Z-line classification. (c) Different images of the same dyed lifted polyps classification.}
\label{fig:attn}
\end{figure}

After the images are preprocessed and grouped into batches of FSL tasks, they are augmented by cross-task interpolation. Randomly select a layer $l$ in the feature extraction process, and representations of the original query and support sets are replaced by the interpolated ones. These encodings are then input to the SCA module, which applies a channel-wise Hadamard product and convolutions to produce self-correlational representations $F_q$ and $F_s$. The representations are processed by the CCA module into a 4-D tensor, from which we constructed affinity graphs for both query and support. The affinity graphs generated by our model are selectively pruned to identify the top-$k$ routes, with resulting attentions illustrated in Fig.\ref{fig:attn}. 

Following the application of the two relational embedding modules, SCA and CCA, our model demonstrates a heightened capability to discern significant regions both within and across images. This proficiency is evident when the model processes pairs of images that either represent zoomed views of the same subject or belong to the same category but depict different subjects. In each case, the model adeptly navigates the classification task. 

\subsection{Comparison to the state-of-the-art methods}
We compared our model to some state-of-the-art as well as several relevant methods on the endoscopic images. Quantitative results are shown in Table.\ref{table:results} with metrics such as Accuracy, Precision, Recall, and F1-score.

\begin{table}[htbp]
\caption{Classification performance on our proposed model and other models}
\begin{center}
\begin{tabularx}{0.4\textwidth}{c c c c c}
\toprule
 &\multicolumn{4}{c}{\textbf{Metrics}} \\
\cline{2-5} 
\textbf{Methods} & \textit{Accuracy}& \textit{Precision}& \textit{Recall} & \textit{F1}\\
\midrule
 MAML \cite{finn_model-agnostic_2017} & 0.792 & 0.610 & 0.633 & 0.621 \\
\midrule
 ProtoNet \cite{snell_prototypical_2017} & 0.775 & 0.662 & 0.694 & 0.678 \\
 \midrule
 Transformer \cite{vaswani_attention_2023} & 0.870 & 0.738 & 0.812 & 0.773 \\
 \midrule
 ResNet50 \cite{resnet} & 0.812 & 0.701 & 0.794 & 0.745 \\
 \midrule
 \textbf{Ours} &\textbf{0.901} & \textbf{0.845} &\textbf{0.942} & \textbf{0.891} \\
\bottomrule
\multicolumn{5}{l}{}
\end{tabularx}
\label{table:results}
\end{center}
\end{table}
 
 Our proposed model demonstrates commendable performance, achieving an accuracy of 90.1\% and an F-score of 0.89 on the dataset. This performance slightly surpasses that of Transformer and is 11\% better than ProtoNet using Conv4.

\section{Ablation}

To investigate the core effect of our main tenets, we conducted extensive ablation studies by omitting or substituting them with existing relevant methods. The different combinations are shown in Table.\ref{tab:abla}, demonstrating the effectiveness of the components in our proposed model. Additionally, we explored the influence of different pretraining datasets on our model's performance. 

\begin{table}[htbp]
    \caption{Results of ablation experiments}
    \centering
    \begin{tabularx}{0.47\textwidth}{X c c}  
        \toprule
        \textbf{Model} & \textbf{\textit{Accuracy}} & \textbf{\textit{Inference Time}} \\
        \midrule
        Ours          &    \textbf{0.901}    &     0.52ms       \\
        \midrule
        Pretrained without ISIC 2018       &    0.869    &     0.53ms       \\        
        \midrule
        Pretrained without Cholec80         &    0.873    &     0.53ms       \\
        \midrule
        Pretrained without Mini-ImageNet          &    0.894    &     0.52ms       \\        
        \midrule
        Using ResNet12 instead of Conv4         &    0.844      &      0.54ms\\
      \midrule
        Keeping Con4 64 layer channels and change relational layer to 64       &  0.898       &       0.52ms\\
        \midrule
        Using MixUp \cite{zhang2017mixup} instead of task interpolation         &    0.878      &      0.45ms\\
      \midrule
        Without task interpolation      &  0.870       &       0.43ms\\
        \midrule
        Using vanilla attention instead of bi-level routing attention            &   0.895    &       0.61ms\\
        \midrule
        Without attention &  0.863 & 0.42ms \\
        \midrule
        Without self correlational representation & 0.886 & 0.52ms \\
        \midrule
        Without cross correlational representation & 0.857 & \textbf{0.41ms}\\

        \bottomrule
    \end{tabularx}
    \label{tab:abla}
\end{table}
Results reported in Table.\ref{tab:abla} reveals following observations:
\begin{itemize}
    \item Omitting ISIC 2018, Cholec80, or Mini-ImageNet during pretraining lowers performance, underscoring the importance of diverse, domain-specific data for robust feature learning and generalization.

    \item Utilizing Conv4 as opposed to ResNet12 and adjusting the relational layer's channel size while maintaining Conv4's 64-layer channels subtly impacts accuracy (0.901 vs. 0.844 and 0.898, respectively). This underscores the significance of the model architecture and relational layer channel dimensions in effectively extracting and relating features from endoscopic images, while also ensuring computational efficiency.
    
    \item Utilizing task interpolation, as opposed to MixUp or no interpolation (accuracy dips to 0.878 and 0.870, respectively), accentuates its role in bolstering model robustness by simulating varied viewpoints, vital for diverse endoscopic image scenarios.
    
    \item The implementation of bi-level routing attention, as opposed to vanilla or no attention (accuracy of 0.895 and 0.863, respectively), demonstrates its proficiency in effectively concentrating on pertinent features in GI images, thereby enhancing classification accuracy.
    
    \item  Excluding self- or cross-correlational representation results in accuracy reductions to 0.886 and 0.857, respectively, highlighting their importance in capturing intra- and inter-image relationships, which is crucial for discerning subtle variations in endoscopic images.
\end{itemize}

\section{Conclusion}
Our DL model excels in analyzing endoscopic images with limited data, using FSL to quickly identify underlying patterns. It combats overfitting through task interpolation, simulating varied camera perspectives for better generalization.

The model's core strength is its ability to capture intra-image details and inter-image connections via self- and cross-correlational embedding. We enhanced it with a bi-level routing attention mechanism, making it lightweight yet focused in image analysis. Trained on diverse datasets including Hyper-kvasir and Mini-ImageNet, it achieved outstanding results on the Kvasir dataset, with 90.1\% accuracy, 0.845 precision, 0.942 recall, and an F1 score of 0.891. This approach marks a significant step forward in endoscopic image analysis. 

\section{Acknowledgment}
This project is supported by the Lee Kong Chian School of Medicine - Ministry of Education Start-Up Grant.


\printbibliography 

@misc{huang2018densely,
      title={Densely Connected Convolutional Networks}, 
      author={Gao Huang and Zhuang Liu and Laurens van der Maaten and Kilian Q. Weinberger},
      year={2018},
      eprint={1608.06993},
      archivePrefix={arXiv},
      primaryClass={cs.CV}
}

@misc{chollet2017xception,
      title={Xception: Deep Learning with Depthwise Separable Convolutions}, 
      author={François Chollet},
      year={2017},
      eprint={1610.02357},
      archivePrefix={arXiv},
      primaryClass={cs.CV}
}

@misc{twinanda2016endonet,
      title={EndoNet: A Deep Architecture for Recognition Tasks on Laparoscopic Videos}, 
      author={Andru P. Twinanda and Sherif Shehata and Didier Mutter and Jacques Marescaux and Michel de Mathelin and Nicolas Padoy},
      year={2016},
      eprint={1602.03012},
      archivePrefix={arXiv},
      primaryClass={cs.CV}
}

@inproceedings{harris_combined_1988,
	title = {A {Combined} {Corner} and {Edge} {Detector}},
	booktitle = {Procedings of the {Alvey} {Vision} {Conference} 1988},
	publisher = {Alvey Vision Club},
	author = {Harris, C. and Stephens, M.},
	year = {1988},
	pages = {23.1--23.6},
}

@article{borgli_hyperkvasir_2020,
	title = {{HyperKvasir}, a comprehensive multi-class image and video dataset for gastrointestinal endoscopy},
	volume = {7},
	copyright = {2020 The Author(s)},
	number = {1},
	journal = {Scientific Data},
	author = {Borgli, Hanna and Thambawita, Vajira and Smedsrud, Pia H. and Hicks, Steven and Jha, Debesh and Eskeland, Sigrun L. and Randel, Kristin Ranheim and Pogorelov, Konstantin and Lux, Mathias and Nguyen, Duc Tien Dang and Johansen, Dag and Griwodz, Carsten and Stensland, Håkon K. and Garcia-Ceja, Enrique and Schmidt, Peter T. and Hammer, Hugo L. and Riegler, Michael A. and Halvorsen, Pål and de Lange, Thomas},
	month = aug,
	year = {2020},
	note = {Number: 1
Publisher: Nature Publishing Group},
	pages = {283},
}

@misc{zhu_biformer_2023,
	title = {{BiFormer}: {Vision} {Transformer} with {Bi}-{Level} {Routing} {Attention}},
	shorttitle = {{BiFormer}},
	publisher = {arXiv},
	author = {Zhu, Lei and Wang, Xinjiang and Ke, Zhanghan and Zhang, Wayne and Lau, Rynson},
	month = mar,
	year = {2023},
	note = {arXiv:2303.08810 [cs]},
}

@misc{vaswani_attention_2023,
	title = {Attention {Is} {All} {You} {Need}},
	publisher = {arXiv},
	author = {Vaswani, Ashish and Shazeer, Noam and Parmar, Niki and Uszkoreit, Jakob and Jones, Llion and Gomez, Aidan N. and Kaiser, Lukasz and Polosukhin, Illia},
	month = aug,
	year = {2023},
	note = {arXiv:1706.03762 [cs]},
}

@inproceedings{dutta_efficient_2021,
  title={Efficient detection of lesions during endoscopy},
  author={Dutta, Amartya and Bhattacharjee, Rajat Kanti and Barbhuiya, Ferdous Ahmed},
  booktitle={Pattern Recognition. ICPR International Workshops and Challenges: Virtual Event, January 10-15, 2021, Proceedings, Part VIII},
  pages={315--322},
  year={2021},
  organization={Springer}
}

@article{ramamurthy_novel_2022,
	title = {A Novel Multi-Feature Fusion Method for Classification of Gastrointestinal Diseases Using Endoscopy Images},
	volume = {12},
	rights = {http://creativecommons.org/licenses/by/3.0/},
	pages = {2316},
	number = {10},
	journaltitle = {Diagnostics},
	author = {Ramamurthy, Karthik and George, Timothy Thomas and Shah, Yash and Sasidhar, Parasa},
}

@misc{finn_model-agnostic_2017,
	title = {Model-Agnostic Meta-Learning for Fast Adaptation of Deep Networks},
	number = {{arXiv}:1703.03400},
	publisher = {{arXiv}},
	author = {Finn, Chelsea and Abbeel, Pieter and Levine, Sergey},
	date = {2017-07-18},
	eprinttype = {arxiv},
	eprint = {1703.03400 [cs]},
}

@inproceedings{snell_prototypical_2017,
	title = {Prototypical Networks for Few-shot Learning},
	volume = {30},
	booktitle = {Advances in Neural Information Processing Systems},
	publisher = {Curran Associates, Inc.},
	author = {Snell, Jake and Swersky, Kevin and Zemel, Richard},
	date = {2017},
}

@misc{yao_meta-learning_2022,
	title = {Meta-Learning with Fewer Tasks through Task Interpolation},
	number = {{arXiv}:2106.02695},
	publisher = {{arXiv}},
	author = {Yao, Huaxiu and Zhang, Linjun and Finn, Chelsea},
	date = {2022-03-17},
	eprinttype = {arxiv},
	eprint = {2106.02695 [cs]},
}

@inproceedings{kvasir,
  title = {KVASIR: A Multi-Class Image Dataset for Computer Aided Gastrointestinal Disease Detection},
  author = {
     Pogorelov, Konstantin and Randel, Kristin Ranheim and Griwodz, Carsten and
     Eskeland, Sigrun Losada and de Lange, Thomas and Johansen, Dag and
     Spampinato, Concetto and Dang-Nguyen, Duc-Tien and Lux, Mathias and
     Schmidt, Peter Thelin and Riegler, Michael and Halvorsen, P{\aa}l
  },
  booktitle = {Proceedings of the 8th ACM on Multimedia Systems Conference},
  year = {2017},
  pages = {164--169},
  publisher = {ACM},
}

@misc{simonyan,
      title={Very Deep Convolutional Networks for Large-Scale Image Recognition}, 
      author={Karen Simonyan and Andrew Zisserman},
      year={2015},
      eprint={1409.1556},
      archivePrefix={arXiv},
      primaryClass={cs.CV}
}

@misc{resnet,
      title={Deep Residual Learning for Image Recognition}, 
      author={Kaiming He and Xiangyu Zhang and Shaoqing Ren and Jian Sun},
      year={2015},
      eprint={1512.03385},
      archivePrefix={arXiv},
      primaryClass={cs.CV}
}

@misc{renet,
      title={Relational Embedding for Few-Shot Classification}, 
      author={Dahyun Kang and Heeseung Kwon and Juhong Min and Minsu Cho},
      year={2021},
      eprint={2108.09666},
      archivePrefix={arXiv},
      primaryClass={cs.CV}
}

@article{imagenet,
Author = {Olga Russakovsky and Jia Deng and Hao Su and Jonathan Krause and Sanjeev Satheesh and Sean Ma and Zhiheng Huang and Andrej Karpathy and Aditya Khosla and Michael Bernstein and Alexander C. Berg and Li Fei-Fei},
Title = {{ImageNet Large Scale Visual Recognition Challenge}},
Year = {2015},
journal   = {International Journal of Computer Vision (IJCV)},
volume={115},
number={3},
pages={211-252}
}

@article{roy_automatic_2022,
  title={Automatic detection and segmentation of colorectal cancer with deep residual convolutional neural network},
  author={Akilandeswari, A and Sungeetha, D and Joseph, Christeena and Thaiyalnayaki, K and Baskaran, K and Jothi Ramalingam, R and Al-Lohedan, Hamad and Al-Dhayan, Dhaifallah M and Karnan, Muthusamy and Meansbo Hadish, Kibrom and others},
  journal={Evidence-Based Complementary and Alternative Medicine},
  volume={2022},
  year={2022},
  publisher={Hindawi}
}

@article{surf,
  title={Surf: Speeded up robust features},
  author={Bay, Herbert and Tuytelaars, Tinne and Van Gool, Luc},
  journal={Lecture notes in computer science},
  volume={3951},
  pages={404--417},
  year={2006},
  publisher={Springer}
}

@inproceedings{orb,
  title={ORB: An efficient alternative to SIFT or SURF},
  author={Rublee, Ethan and Rabaud, Vincent and Konolige, Kurt and Bradski, Gary},
  booktitle={2011 International conference on computer vision},
  pages={2564--2571},
  year={2011},
  organization={Ieee}
}

@article{codella2019skin,
  title={Skin lesion analysis toward melanoma detection 2018: A challenge hosted by the international skin imaging collaboration (isic)},
  author={Codella, Noel and Rotemberg, Veronica and Tschandl, Philipp and Celebi, M Emre and Dusza, Stephen and Gutman, David and Helba, Brian and Kalloo, Aadi and Liopyris, Konstantinos and Marchetti, Michael and others},
  journal={arXiv preprint arXiv:1902.03368},
  year={2019}
}

@article{tschandl2018ham10000,
  title={The HAM10000 dataset, a large collection of multi-source dermatoscopic images of common pigmented skin lesions},
  author={Tschandl, Philipp and Rosendahl, Cliff and Kittler, Harald},
  journal={Scientific data},
  volume={5},
  number={1},
  pages={1--9},
  year={2018},
  publisher={Nature Publishing Group}
}

@article{lu2022real,
  title={Real-time automated diagnosis of colorectal cancer invasion depth using a deep learning model with multimodal data (with video)},
  author={Lu, Zihua and Xu, Youming and Yao, Liwen and Zhou, Wei and Gong, Wei and Yang, Genhua and Guo, Mingwen and Zhang, Beiping and Huang, Xu and He, Chunping and others},
  journal={Gastrointestinal Endoscopy},
  volume={95},
  number={6},
  pages={1186--1194},
  year={2022},
  publisher={Elsevier}
}

@article{luo_diagnosis_2022,
  title={Diagnosis of ulcerative colitis from endoscopic images based on deep learning},
  author={Luo, Xudong and Zhang, Junhua and Li, Zonggui and Yang, Ruiqi},
  journal={Biomedical Signal Processing and Control},
  volume={73},
  pages={103443},
  year={2022},
  publisher={Elsevier}
}

@article{zeng_image_2021,
  title={An image classification model based on transfer learning for ulcerative proctitis},
  author={Zeng, Feng and Li, Xingcun and Deng, Xiaoheng and Yao, Lan and Lian, Guanghui},
  journal={Multimedia Systems},
  pages={1--10},
  year={2021},
  publisher={Springer}
}

@article{zhang2017mixup,
  title={mixup: Beyond empirical risk minimization},
  author={Zhang, Hongyi and Cisse, Moustapha and Dauphin, Yann N and Lopez-Paz, David},
  journal={arXiv preprint arXiv:1710.09412},
  year={2017}
}

\end{document}